\documentclass[runningheads]{llncs}
\usepackage[T1]{fontenc}
\usepackage{graphicx}
\usepackage{amsmath}
\usepackage{amssymb}
\usepackage{booktabs}
\usepackage{amsfonts}       
\usepackage{xcolor}         
\usepackage{color, colortbl}

\usepackage{adjustbox}
\usepackage{multicol}
\usepackage{multirow}
\usepackage{units}
\usepackage{pifont}
\usepackage{dsfont}
\usepackage{subcaption}
\usepackage{wrapfig}
\usepackage{pifont}
\usepackage{threeparttable}

\usepackage[pagebackref,breaklinks,colorlinks]{hyperref}

\begin{document}
\title{HOI-R1: Exploring the Potential of Multimodal Large Language Models for Human-Object Interaction Detection}
\titlerunning{HOI-R1: Exploring the Potential of MLLMs for HOID}
%
\author{Junwen Chen \and
Peilin Xiong \and
Keiji Yanai}
\authorrunning{Junwen Chen, Peilin Xiong, and Keiji Yanai}
%
\institute{Department of Informatics, The University of Electro-Communications, Tokyo, Japan
\email{\{chen-j, xiong-p, yanai\}@mm.inf.uec.ac.jp}}
\maketitle              
\begin{abstract}
Recent human-object interaction detection (HOID) methods highly require prior knowledge from vision-language models (VLMs) to enhance the interaction recognition capabilities. The training strategies and model architectures for connecting the knowledge from VLMs to the HOI instance representations from the object detector are challenging, and the whole framework is complex for further development or application. On the other hand, the inherent reasoning abilities of multimodal large language models (MLLMs) on human-object interaction detection are under-explored. Inspired by the recent success of training MLLMs with reinforcement learning (RL) methods, we propose HOI-R1 and first explore the potential of the language model on the HOID task without any additional detection modules. We introduce an HOI reasoning process and HOID reward functions to solve the HOID task by pure text.
Experiments on HICO-DET across multiple open-source MLLMs, including the Qwen-VL family (Qwen2.5-VL and Qwen3-VL) and Rex-Omni, show consistent improvements. Especially, HOI-R1 boosts Qwen2.5-VL-3B 2$\times$ accuracy with great generalization ability.
The source code is available at \url{https://github.com/cjw2021/HOI-R1}.

\end{abstract}

\section{Introduction}
\label{sec:intro}

Human–Object Interaction Detection (HOID) aims to detect interactions between humans and surrounding objects in an image. Given an image, the goal is to predict a set of HOI instances represented as \{$B_h$, $B_o$, \textit{Object Class}, \textit{Interaction Class}\}. Beyond recognizing the relationship category, HOID requires precise localization and pairing of human–object regions, making it a challenging form of fine-grained relationship understanding with grounding.

\begin{figure}[!t]
    \centering
    \includegraphics[width=0.7\linewidth]{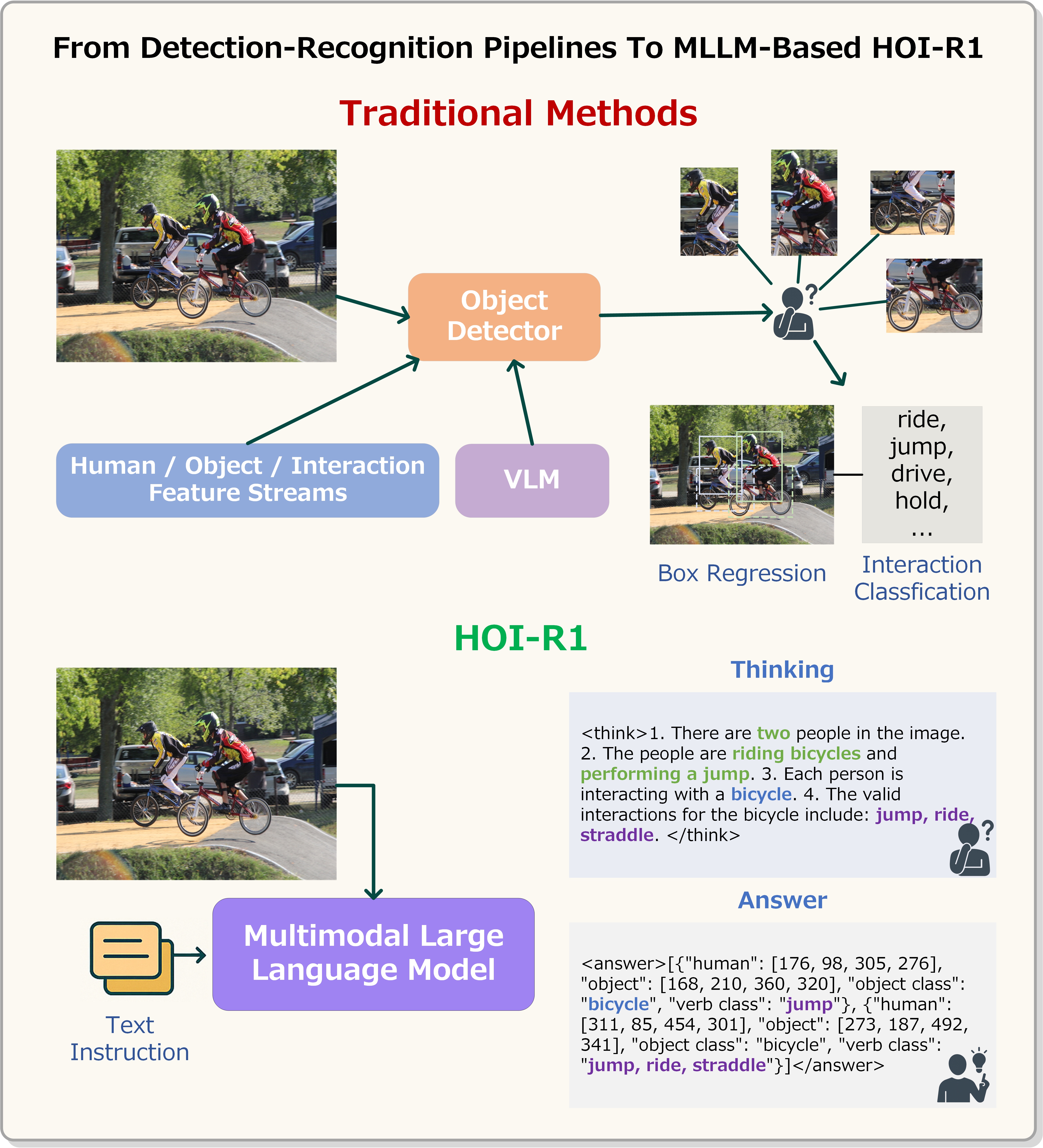}
    \caption{Comparison of the pipeline of traditional HOID methods and our proposed HOI-R1. Traditional HOID methods rely on object detectors to extract HOI embeddings, while HOI-R1 directly interprets interactions through natural language reasoning using MLLMs.}
    \label{fig:framework_comparison}
\end{figure}

Most existing HOID approaches~\cite{tamura2021qpic,zou2021end,kim2021hotr} follow a detector-centric paradigm: an object detector provides human/object proposals, and a dedicated interaction head (often transformer-based) predicts HOI labels. Due to long-tailed HOI annotations, recent state-of-the-art methods~\cite{ning2023hoiclip,cao2024detecting,chen2025focusing} increasingly rely on VLM priors (e.g., CLIP~\cite{radford2021learning}/BLIP~\cite{li2023blip}-style representations) to enhance interaction recognition. However, transferring such priors typically introduces additional modules, training objectives, and architectural coupling between detectors and interaction decoders, leading to complex pipelines that are hard to extend or reuse.

In contrast, modern multimodal large language models (MLLMs)~\cite{li2024llava,bai2025qwen2,zhu2025internvl3} have shown strong capabilities in fine-grained visual perception, spatial grounding, and multimodal reasoning. In particular, the Qwen-VL family (Qwen2.5-VL~\cite{bai2025qwen2} and the more recent Qwen3-VL~\cite{Qwen3-VL}) provides general-purpose grounded generation, while Rex-Omni~\cite{jiang2025rexomni} reformulates visual perception tasks as next-point prediction and exhibits strong localization priors.
DeepSeek-R1~\cite{guo2025deepseek} exemplifies this trend, demonstrating that RL can induce powerful reasoning behaviors even without supervised fine-tuning (SFT) as a preliminary step.
Recent studies~\cite{huang2025vision,tan2025reason,shen2025vlm} also validate RL's effectiveness for aligning MLLMs with visual reasoning tasks.
Despite these advances, the capability of MLLMs for structured HOID—simultaneously reasoning about interactions and localizing HOI pairs in a standardized evaluation protocol—remains underexplored.

\begin{figure}[!t]
    \centering
    \includegraphics[width=0.65\linewidth]{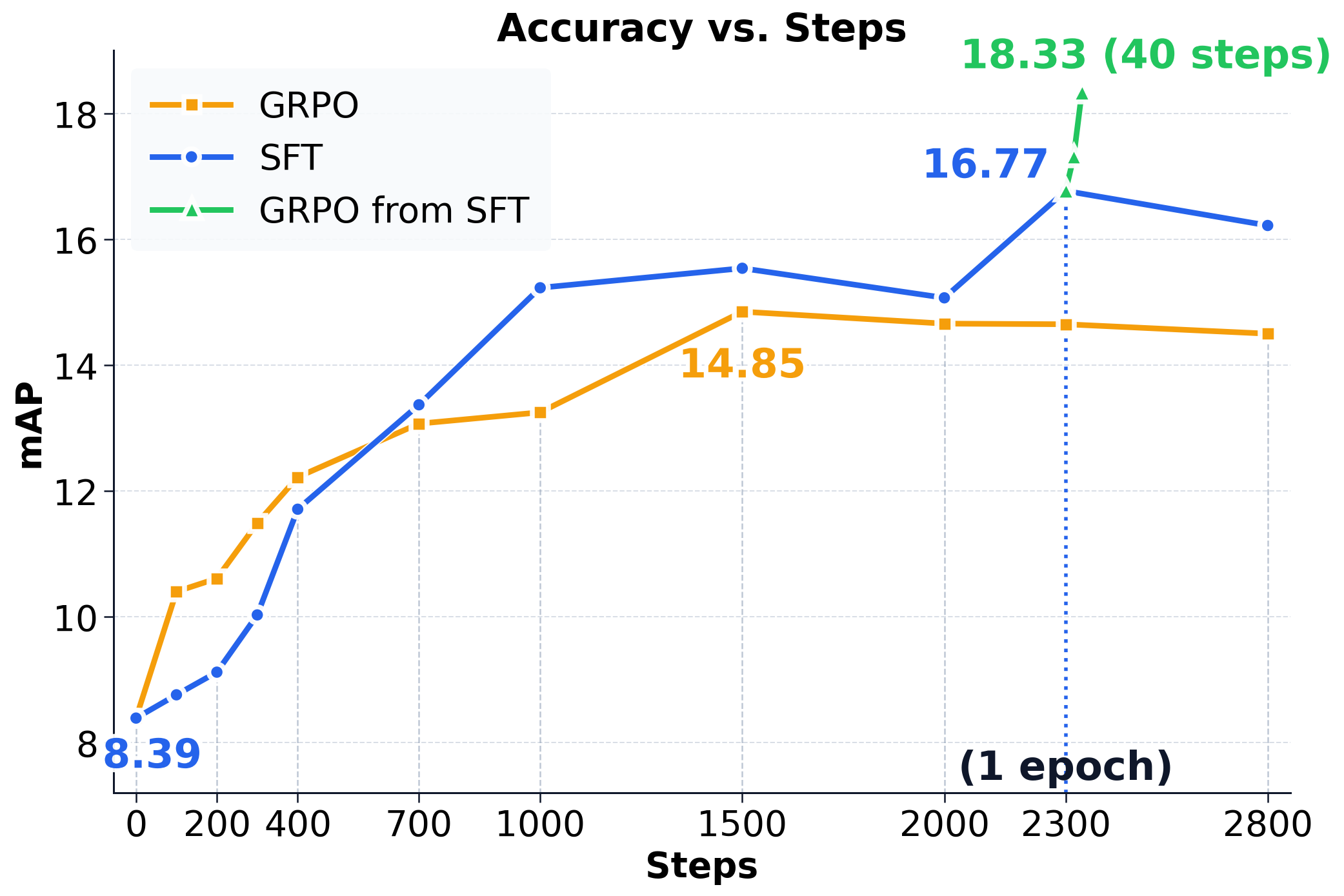}
    \caption{Training convergence of HOI-R1 with Qwen2.5-VL-3B on HICO-DET. The mAP of \textit{Full} category on \textit{Default Setting} is shown. HOI-R1 achieves more than 2x performance boost with only 1 epoch SFT and 40 steps RL training.}
    \label{fig:results_all}
\end{figure}

To this end, we first explore their tremendous potential in HOID tasks, as shown in Fig.~\ref{fig:framework_comparison}. We propose \textbf{HOI-R1}, a radical shift: replacing detectors with natural language reasoning, using MLLMs to directly interpret interactions through holistic scene understanding in both visual and textual modalities.
Solving HOID purely through natural language reasoning requires simultaneous prediction of multiple bounding boxes, precise pairing of objects with their interactions, and accurate relationship recognition—all within a complex, structured reasoning pipeline. In practice, we design a systematic prompt structure to guide the reasoning, which injects HOI knowledge through SFT with thinking distillation.
As shown in Fig.~\ref{fig:results_all}, with HOI knowledge distillation, MLLM shows a significant performance boost.
Then, we introduce RL for further alignment with four reward functions, including format rewards for output structure, object/interaction label accuracy, and a one-to-one matching HOI IoU reward, and the performance can be improved with only 100 training steps.
\begin{itemize}
    \item We introduce HOI-R1, the first MLLM framework that solves HOID end-to-end via natural language, eliminating object detectors.
    \item We introduce an SFT with thinking distillation to extend the HOI knowledge and a reinforcement learning (RL) paradigm to align the MLLM on HOID with our HOI reward functions to further enhance the performance.
    \item We demonstrate consistent gains across multiple open-source MLLMs on a standard HOID benchmark, establishing new baselines for MLLM-based HOID.
\end{itemize}

\section{Related Work}
\label{sec:related_work}

\noindent \textbf{MLLMs for vision tasks.} \quad Recent MLLMs have made rapid progress in fine-grained visual understanding and grounded generation. The Qwen-VL family provides strong multimodal perception and spatial grounding for both images and videos \cite{bai2025qwen2,Qwen3-VL}. 
Qwen2.5-VL~\cite{bai2025qwen2} combines dynamic-resolution visual encoding and absolute time-aware video modeling to achieve fine-grained perception, robust structured understanding of complex images and long-horizon videos.
Qwen3-VL~\cite{Qwen3-VL} supports up to 256K-token interleaved multimodal contexts and, through architectural upgrades such as enhanced interleaved-MRoPE, DeepStack-based multi-level ViT fusion, and text-aligned temporal grounding, delivers superior pure-text understanding, long-context multimodal comprehension, and advanced visual–math reasoning.
Rex-Omni~\cite{jiang2025detect} reformulates object detection and related perception tasks as discrete coordinate token prediction, leveraging quantized 0–999 coordinate tokens, achieve state-of-the-art zero-shot object perception while enabling versatile language-aware visual grounding capabilities such as referring, pointing, GUI grounding, and OCR.

\noindent \textbf{RL enhanced MLLMs on Visual Reasoning Tasks.} \quad Vision-R1~\cite{huang2025vision} uses an MLLM that combines cold-start initialization with RL to enhance reasoning capabilities on math benchmarks while generating human-like reasoning processes.
Reason-RFT~\cite{tan2025reason} introduces a novel two-phase reinforcement fine-tuning framework that combines SFT with Chain-of-Thought (CoT) reasoning activation and Group Relative Policy Optimization (GRPO)~\cite{shao2024deepseekmath} to enhance generalization in visual reasoning tasks.
VLM-R1~\cite{shen2025vlm} focuses on tasks like Referring Expression Comprehension (REC) and Open-Vocabulary Object Detection (OVD), leveraging deterministic ground-truth annotations for stable reward computation. The study also highlights key insights such as reward hacking in object detection and the emergence of the "OD aha moment," where models first reason about object presence before localization.
Based on these successful experiences, we extend the capabilities of MLLM to the more complex HOI task, design input instructions, training strategies, and reward functions for HOID, and provide new possible directions for further development in this field.

\begin{figure}[!t]
    \centering
    \includegraphics[width=\linewidth]{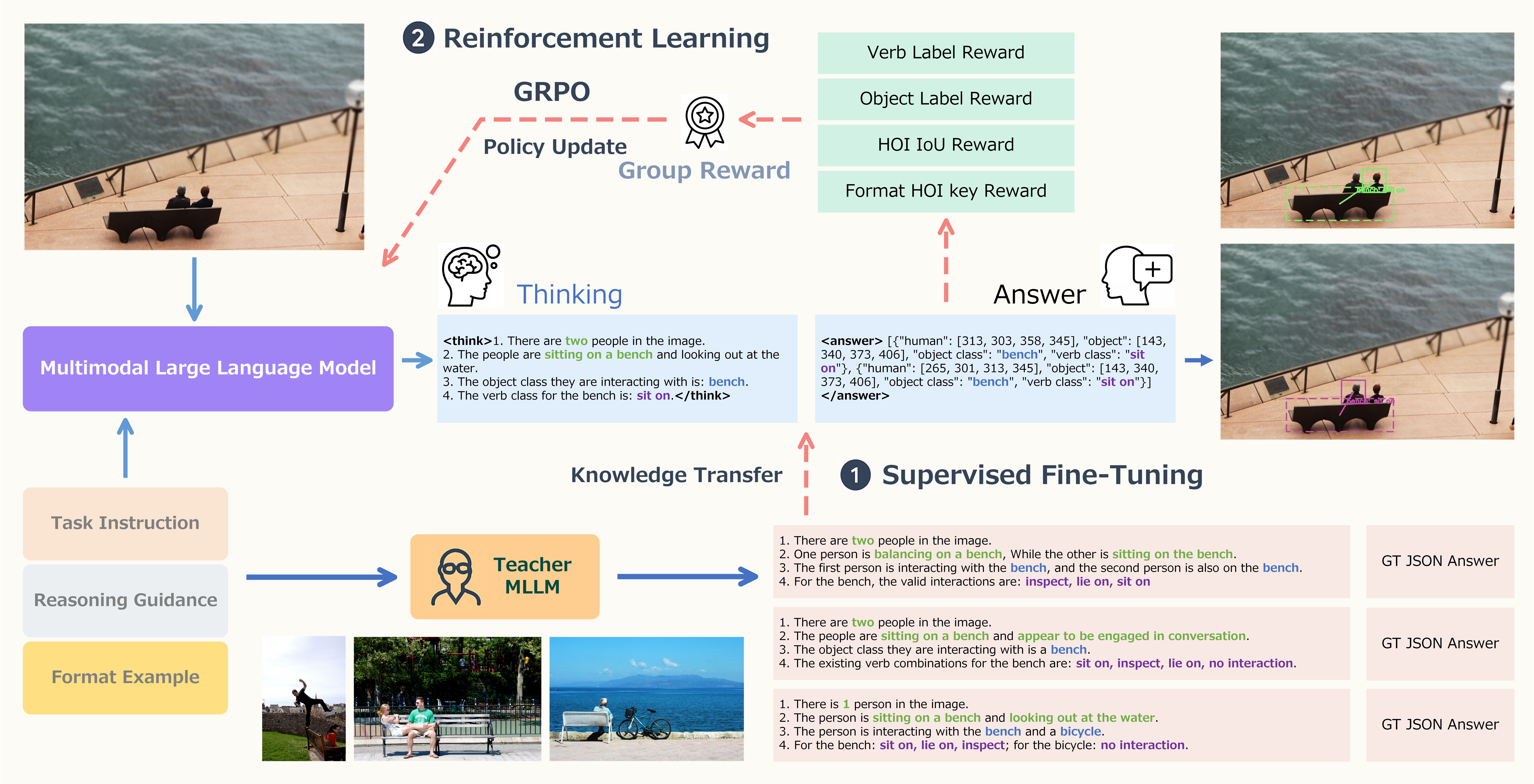}
    \caption{Overview of our HOI-R1 framework. The input consists of two modalities: image and text. The question text consists of three parts: the task instruction includes basic information about the task, the reasoning guidance provides hints for the reasoning process, and the format example regularizes the output. First, a Teacher MLLM model is used to generate reasoning steps for Supervised Fine-tuning (SFT). Then, in the Reinforcement Learning (RL) stage, the student MLLM model, as the policy model, is trained with four reward signals.}
    \label{fig:arch}
\end{figure}

\section{Metholodology}
\label{sec:framework}

\subsection{Language-based HOID Prediction}
\label{sec:hoi_prediction_definition}
\begin{figure}[!t]
    \centering
    \includegraphics[width=0.6\linewidth]{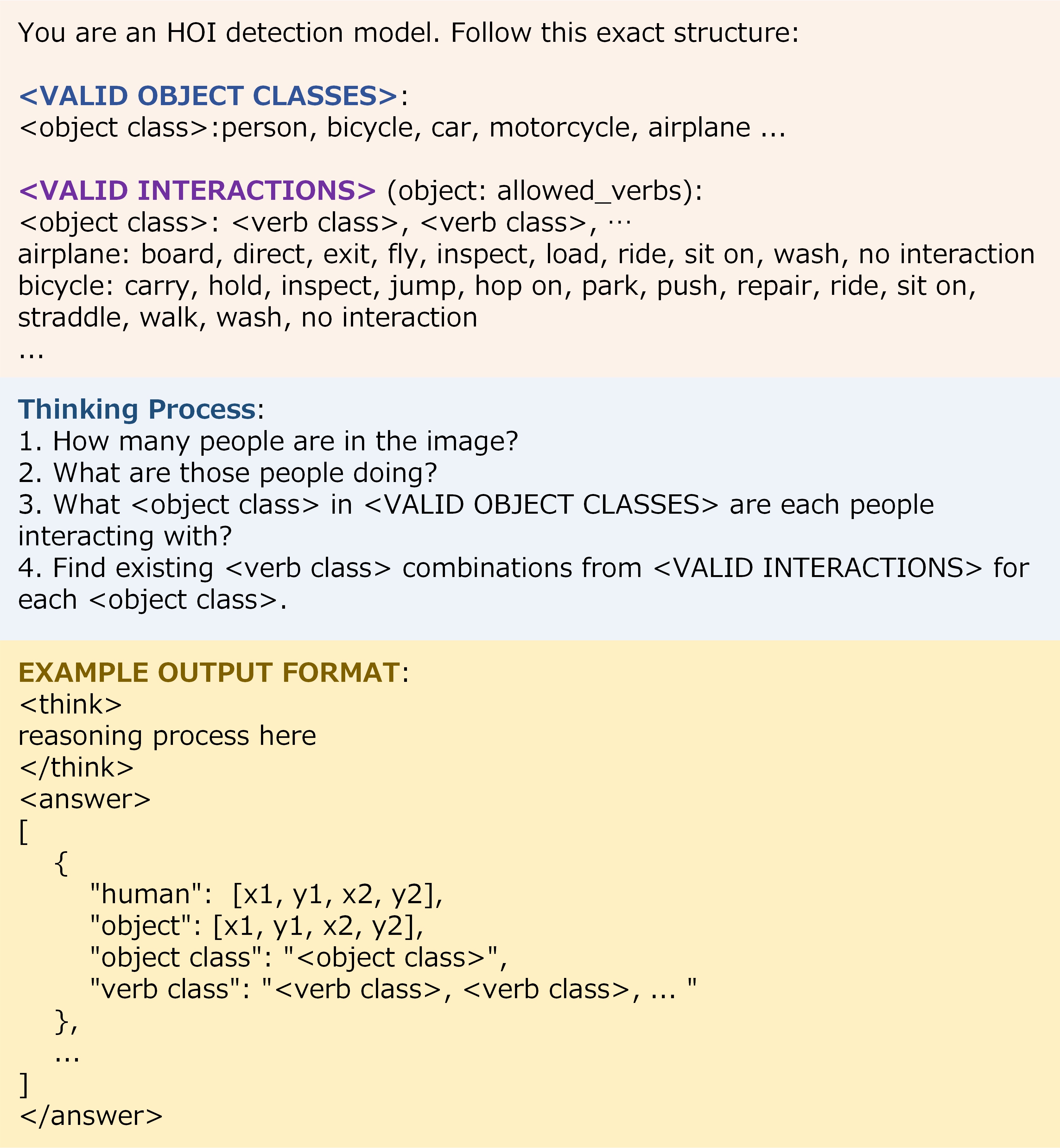}
    \caption{The input question template for HOI-R1. The template consists of three key components: Task Instruction, Reasoning Guidance, and Format Example.}
    \label{fig:question_template}
\end{figure}

As shown in Fig.~\ref{fig:framework_comparison}, unlike conventional HOI detection methods that rely on bounding box regression and interaction classification, we propose a novel language-based paradigm that directly outputs all HOI instances in natural language format. Without modifying the model architecture or compromising its original capabilities, we design the input question template to effectively elicit the model's HOI detection potential. We illustrate the detailed design of the question template in Fig.~\ref{fig:question_template}. The question template consists of three key components to guide the model:

\noindent \textbf{Task Instruction}: We first establish the model's role ("You are an HOI detection model") and provide the complete vocabulary for both objects and interactions in the HOID dataset. The exhaustive list of \texttt{<VALID OBJECT CLASSES>} and \texttt{<VALID INTERACTIONS>} serves as a constrained output space, ensuring the model's predictions align with standard HOI benchmarks while preventing hallucination of irrelevant categories.

\noindent \textbf{Reasoning Guidance}: The "Thinking Process" breaks down the complex HOI detection task into sequential reasoning steps, mirroring human cognitive processes. First, the MLLM is required to identify humans in the scene, then analyze their actions, and finally determine their interactions with surrounding objects. This step-by-step decomposition enables the model to handle the compositional nature of HOI relationships systematically.

\noindent \textbf{Format Example}: The output template demonstrates the expected JSON structure containing both the reasoning chain (\texttt{<think>} tag) and final HOI predictions (\texttt{<answer>} tag). As recent MLLMs are trained to represent bounding boxes, we directly incorporate the spatial coordinates into the language output.

The template design offers several advantages: (1) The explicit verb-object compatibility list (\texttt{<VALID INTERACTIONS>}) avoids hallucination of unlikely verb-object combinations; (2) The thinking process prompts enable the model to leverage its inherent reasoning capabilities for complex scene understanding; (3) The structured output format bridges the gap between free-form language generation and standardized HOI detection requirements.

\subsection{Thinking Distillation via SFT}
\label{sec:sft}

To transfer task-specific knowledge to the student MLLM, we employ a supervised fine-tuning (SFT) stage with a teacher-student knowledge transfer paradigm. This process, termed \textit{thinking distillation}, leverages a powerful teacher model to generate reasoning traces that guide the student model's learning process.

\noindent \textbf{Teacher Reasoning Generation:} 
We utilize GPT4o-mini~\cite{achiam2023gpt} as the teacher model to generate step-by-step reasoning traces. For each image in the training set of HICO-DET, we input the image along with the structured prompt's \textit{Reasoning Guidance} (as defined in Section~\ref{sec:hoi_prediction_definition}). The teacher model produces natural language reasoning sequences enclosed within \texttt{<think>} tags.
These distilled reasoning traces capture the implicit logical process of HOI detection that traditional supervised learning fails to explicitly teach.

\noindent \textbf{Thinking Distillation and Answer Supervision:}
With the ground-truth annotations from the dataset, we supervise the student model to learn both the reasoning traces and the final HOI predictions. The student model is trained to predict two components: (1) predicting the teacher-generated \texttt{<think>} sequences, internalizing the step-by-step HOI reasoning logic. (2) \texttt{<answer>} component is directly supervised using ground-truth HOI triplets from HICO-DET annotations. This ensures precise alignment with the target task objectives while maintaining output fidelity.
Given image $x$, question template $q$, teacher-generated reasoning $r$, and ground-truth answer $a$, the training objective is the autoregressive negative log-likelihood over the supervised tokens:
\begin{equation}
\begin{aligned}
\mathcal{L}_{\text{SFT}}
= -\,\mathbb{E}_{(x,q,r,a)\sim\mathcal{D}} \Bigg[
\underbrace{\sum_{t=1}^{T_r}\log \pi_\theta\!\big(r_t \mid x,q,r_{<t}\big)}_{\text{\small reasoning supervision}} +
\underbrace{\sum_{t=1}^{T_a}\log \pi_\theta\!\big(a_t \mid x,q,r,a_{<t}\big)}_{\text{\small answer supervision}}
\Bigg]
\end{aligned}
\end{equation}
where $\pi_\theta$ is the student MLLM parameterized by $\theta$, and $\mathcal{D}$ is the training dataset.
As shown in Fig.~\ref{fig:results_all}, the SFT stage establishes a strong foundation for the subsequent reinforcement learning alignment, which further refines the model's outputs using HOID-specific rewards.

\begin{figure}[!t]
    \centering
    \includegraphics[width=0.95\linewidth]{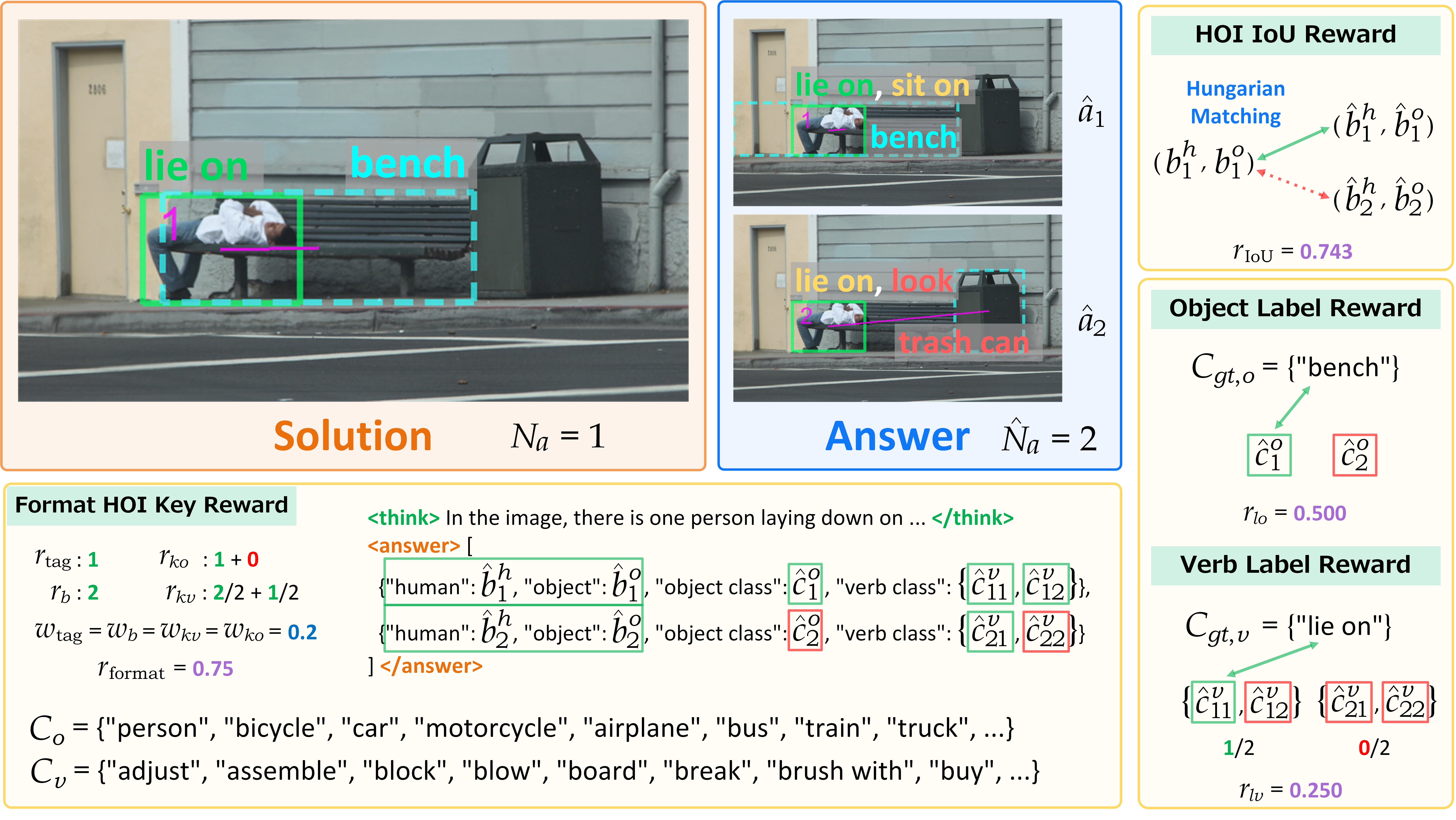}
    \caption{The reward functions of HOI-R1. We design key format reward, label reward, and label reward to ensure the structural, semantic, and geometric alignment of the model outputs with the ground truth.}
    \label{fig:reward_example}
\end{figure}

\subsection{HOID Reinforcement Learning}
\label{sec:hoi_rl}
After establishing foundational capabilities via SFT, we further align the student MLLM through Reinforcement Learning (RL) to enforce structural, semantic, and geometric alignment with ground truth.
Following recent successes, we employ the Group Relative Policy Optimization (GRPO)~\cite{shao2024deepseekmath} algorithm, which is efficient for post-training LLMs and MLLMs.
For each input image $x$ with question template $q$, GRPO samples $G$ outputs $\{o_i\}^G_{i=1}$ from the old policy $\pi_{\theta_{old}}$ and optimizes the policy with the following objective:
\begin{equation}
    \begin{aligned}
        \mathcal{J}_{\text{GRPO}}
        & = -\,\mathbb{E}_{(x,q)\sim D, \{o_i\}^G_{i=1}\sim \pi_{\theta_{old}}(O|x,q)}
        \\
        & \frac{1}{G} \sum_{i=1}^{G} \Big\{ \text{min}[s_1\cdot \hat{A}_i, s_2\cdot \hat{A}_i] - \beta\mathbb{D}_{\text{KL}}[\pi_\theta||\pi_{\text{ref}}] \Big\}
    \end{aligned}
\end{equation}
\begin{align}
    & s_1 = \frac{\pi_\theta(o_i|x,q)}{\pi_{\theta_{old}}(o_i|x,q)}
    \\
    & s_2 = \text{clip}(\frac{\pi_\theta(o_i|x,q)}{\pi_{\theta_{old}}(o_i|x,q)}, 1-\epsilon, 1+\epsilon)
    \\
    & \hat{A}_i = \frac{r_i - \text{mean}(\{r_1,r_2,\ldots,r_G\})}{\text{std}(\{r_1,r_2,\ldots,r_G\})}
\end{align}
where $\hat{A}_i$ is the advantage for output $o_i$, and $\mathbb{D}_{\text{KL}}$ is the KL divergence.
The reward function is crucial for guiding the model towards predictions.
As an HOI instance involves multiple elements (e.g., human, object, object class, and verb class), to ensure fine-grained alignment, we design element-specific rewards that guide the learning process more comprehensively. In Fig.~\ref{fig:reward_example}, we illustrate the detailed design of our reward functions, which consist of three components: (1) HOI key format reward, (2) object and verb label reward, and (3) HOI IoU reward. Each component is described in detail below.

\noindent \textbf{HOI Key Format Reward:} Since the predictions are generated in plain text, it is necessary to ensure the correctness of the output format. Therefore, we design a format reward for each key-value pair in the dict of every HOI instance within the prediction list.
First, the model output must contain the \texttt{<answer>} tag; otherwise, all rewards are set to zero. We divide the rewards into five components: the reward for the \texttt{<think>} tag $r_{\text{tag}}$, the reward for human and object boxes $r_{b}$, the reward for object label $r_{ko}$, and the reward for verb label $r_{kv}$. Specifically, for the entire output text, the thinking tag reward is defined as:
\begin{equation}
    r_{\text{tag}} = \mathbf{1}[\texttt{"<think>"} \in \hat{y}]
    \label{eq:tag_reward}
\end{equation}
where $\mathbf{1}[\cdot]$ is the indicator function, and $\hat{y}$ is the model output text.
Then, considering the $i$-th predicted HOI instance dict $\hat{a}_i$ in the answer list, the box format reward $r_{b_i}$ is defined as:
\begin{equation}
    \begin{aligned}
        r_{b_i} & = \mathbf{1}\Big[ \{\texttt{"human"},\texttt{"object"}\} \subseteq \text{keys}(\hat{a}_i) \;\wedge\; 
        \\
        & \forall b \in \tilde{B}_i: \text{IoU}(\hat{b}_i^o, b^o) \leq 0.5 \;\wedge\; \text{IoU}(\hat{b}_i^h, b^h) \leq 0.5 \Big]
    \end{aligned}
    \label{eq:box_reward}
\end{equation}
\begin{equation}
    B_i = B_{i-1} \cup \{(\hat{b}_i^h, \hat{b}_i^o)\}, \quad B_0 = \emptyset
\end{equation}
where $\hat{b}_i^h$ and $\hat{b}_i^o$ are the predicted human and object bounding boxes, respectively, and $\tilde{B}$ is the set of all previously predicted boxes such that both the human and object boxes have IoU less than 0.5 with each other.
Both the keys ``human'' and ``object'' must exist in the dict $\hat{a}_i$, and we defined $\tilde{B}$ that records all unique boxes to avoid reward hacking by duplicated boxes. Moreover, if an instance's boxes are duplicated, no further rewards are computed for that instance.
For the object-label reward $r_{ko_i}$, the key ``object class'' must exist, and its value $\hat{c}_i^o$ must belong to the predefined object-class set $C_a$.
\begin{equation}
    r_{ko_{i}} = \mathbf{1}[\texttt{"object class"} \in \text{keys}(\hat{a}_i) \;\wedge\; \hat{c}_i^o \in C_o]
    \label{eq:object_label_reward}
\end{equation}
Different from the object class, since a single HOI instance may involve multiple interactions, we compute the ratio between the number of distinct labels belonging to the verb-class set $C_v$ and the total number of predicted verb labels as the reward.
\begin{equation}
    r_{kv_i} = \frac{|\{\tilde{c}_i^v\}|}{|\{\hat{c}_{i}^v\}|} \cdot \mathbf{1}[\texttt{"verb class"} \in \text{keys}(\hat{a}_i)]
    \label{eq:verb_label_reward}
\end{equation}
\begin{equation}
    \tilde{c}_i^v = \text{Unique}(\{\hat{c}_{ij}^v \mid \hat{c}_{ij}^v \in C_v, j=1,2,\ldots,|\{\hat{c}_{i}^v\}|\})
\end{equation}
where $\hat{c}_{ij}^v$ is the $j$-th predicted verb label in the $i$-th HOI instance, and $\text{Unique}(\cdot)$ returns the set of unique elements.
In addition, we introduce a key penalty to avoid duplicate keys in the dict of each HOI instance as follows:
\begin{equation}
    \alpha_i = \frac{N_k}{N_k + |\hat{N}_{k_i} - N_k|}
\end{equation}
where $N_k=4$ is the standard number of keys, and $\hat{N}_{k_i}$ is the number of keys in the $i$-th predicted HOI instance.
Finally, all component rewards are combined with weights to form the overall format reward:
\begin{equation}
    r_{\text{format}} = w_{\text{tag}}r_{\text{tag}} + \frac{\sum_{i=1}^{\hat{N}_a} \alpha_i \big[ w_{b}r_{b_i} + w_{ko}r_{ko_i} + w_{kv}r_{kv_i} \big]}{\text{max}(N_a, \hat{N}_a)}
\end{equation}
where $\hat{N}_a$ and $N_a$ are the numbers of predicted and ground-truth HOI instances, respectively, and $w_{tag}$, $w_{b}$, $w_{ko}$, and $w_{kv}$ are the weights for each component.

\noindent \textbf{Object and Verb Label Reward:} Different from the format reward, which is irrelevant to specific labels, the purpose of label reward is to encourage the model to make more accurate predictions of the HOIs that appear in the ground truth.
In practice, we compare the predicted labels in each HOI instance $a_i$ with the ground-truth label set $C_{gt}$ one by one in sequential order with a drop-on-match strategy. The object label rewards is defined as:
\begin{equation}
    r_{lo} = \frac{\sum_{i=1}^{\hat{N}_a} \alpha_i \mathbf{1}[\hat{c}_i^o\in C_{gt,o}^{(i-1)}]}{\text{max}(N_a, \hat{N}_a)}
\end{equation}
where
\begin{equation}
    C_{gt,o}^{(i)} =
    \begin{cases}
        C_{gt,o}^{(i-1)} \setminus \{\hat{c}_i^o\}, & \text{if } \hat{c}_i^o \in C_{gt,o}^{(i-1)}, \\
        C_{gt,o}^{(i-1)}, & \text{otherwise}.
    \end{cases}
\end{equation}
and the verb label reward is defined as:
\begin{equation}
    r_{lv} = \frac{\sum_{i=1}^{\hat{N}_a} \frac{\alpha_i}{|\{\hat{c}_{i}^v\}|} \sum_{j=1}^{|\{\hat{c}_{i}^v\}|} \mathbf{1}[\hat{c}_{ij}^v\in C_{gt,v}^{(i-1)}]}{\text{max}(N_a, \hat{N}_a)}
\end{equation}
where
\begin{align}
    & C_{gt,v}^{(i)} = C_{gt,v}^{(i-1)} \setminus \{\bar{c}_i^v\}
    \\
    & \{\bar{c}_i^v\} = \{\hat{c}_{ij}^v \mid \hat{c}_{ij}^v \in C_{gt,v}^{(i-1)}, j=1,2,\ldots,|\{\hat{c}_{i}^v\}|\}
\end{align}

\noindent \textbf{HOI IoU Reward:} Inspired by recent transformer-based HOID methods~\cite{tamura2021qpic}, we leverage the Hungarian algorithm~\cite{kuhn1955hungarian} to match the predicted HOI boxes and ground-truth boxes for accurate spatial alignment. The cost matrix, considering the Intersection over Union (IoU) of HOI pairs, is defined as:
\begin{equation}
    C_{ij} = 1 - s_{ij}
\end{equation}
\begin{equation}
    s_{ij} = \frac{1}{2} \big[ \text{IoU}(\hat{b}^{h}_{i}, b^{h}_{j}) + \text{IoU}(\hat{b}^{o}_{i}, b^{o}_{j}) \big]
\end{equation}
where $\hat{b}^{h}_{i}$ and $\hat{b}^{o}_{i}$ are the predicted human and object bounding boxes, respectively. The one-to-one matching $\mathcal{M}^{*}$ is obtained by solving the linear assignment problem:
\begin{equation}
    \mathcal{M}^{*} = \text{argmin}_{\mathcal{M}} \sum_{(i,j) \in \mathcal{M}} C_{ij}
\end{equation}
The final reward is defined as:
\begin{equation}
    r_{\text{IoU}} = \frac{1}{N_a} \sum_{(i,j) \in \mathcal{M}^{*}} s_{ij}
\end{equation}
Note that the one-to-one matched predicted HOI instances can not be more than the ground-truth HOI instances, i.e., $|\mathcal{M}^{*}| \leq N_a$, we use $N_a$ is used to normalize the reward.

Finally, the overall reward considering all components is defined as:
\begin{equation}
    r = r_{\text{format}} + r_{lo} + r_{lv} + r_{\text{IoU}}
\end{equation}
With our HOID-specific rewards, the model is effectively guided to produce accurate and well-structured HOI predictions.

\section{Experiments}
\label{sec:experiments}

In this section, we present the experimental setup, datasets, and evaluation metrics used to assess the performance of our proposed method. We also provide a detailed analysis of the results obtained from various experiments conducted to validate our approach.

\begin{table}[!ht]
    \centering
    \resizebox{0.95\linewidth}{!}{
    \begin{tabular}{@{}llccccccc@{}}
        \toprule
        \multirow{2.5}{*}{Method} & \multirow{2.5}{*}{Training Sessions} & \multicolumn{3}{c}{Default} & \multicolumn{3}{c}{Known Object} \\
        \cmidrule(lr){3-5} \cmidrule(lr){6-8}
        & & Full & Rare & Non-Rare & Full & Rare & Non-Rare \\
        \midrule \midrule
        \multicolumn{8}{@{}l}{\textit{--- Traditional HOID Methods ---}} \\
        \addlinespace
        HO-RCNN~\cite{chao2018learning} & 150k & 7.81 & 5.37 & 8.54 & 10.41 & 8.94 & 10.85 \\
        HOTR~\cite{kim2021hotr}         & 100 epoch & 25.10 & 17.34 & 27.42 & - & - & - \\
        HOI-Trans~\cite{zou2021end}     & 150 epoch & 26.61 & 19.15 & 28.84 & 29.13 & 20.98 & 31.57 \\
        QPIC~\cite{tamura2021qpic}      & 150 epoch & 29.07 & 21.85 & 31.23 & 31.68 & 24.14 & 33.93 \\
        \midrule \midrule
        \multicolumn{8}{@{}l}{\textit{--- MLLMs ---}} \\
        \rowcolor{gray!20}
        Qwen2.5-VL-3B (baseline)  & -         & 8.39  & 9.60  & 8.03  & 8.96  & 9.83  & 8.70 \\
        Qwen2.5-VL-7B             & -         & 10.46 & 14.30 & 9.31  & 11.01 & 14.63  & 9.93 \\
        Qwen2.5-VL-32B-AWQ        & -         & 18.12 & 24.56 & 16.20 & 19.90 & 25.77 & 18.15 \\
        Qwen2.5-VL-72B-AWQ        & -         & 20.71 & 29.62 & 18.05 & 22.93 & 31.87 & 20.26 \\
        \rowcolor{gray!20}
        Qwen3-VL-4B (baseline)    & -         & 16.16 & 22.31 & 14.32 & 17.55 & 22.87 & 15.96 \\
        Qwen3-VL-8B               & -         & 17.41 & 21.96 & 16.05 & 19.17 & 23.39 & 17.91 \\
        \midrule \midrule
        \multicolumn{8}{@{}l}{\textit{--- Supervised Fine-Tuning (SFT) or Reinforcement Learning (RL) ---}} \\
        \addlinespace
        Qwen2.5-VL-3B-SFT         & 400 steps & 11.71 & 10.52 & 12.07 & 12.60 & 10.70 & 13.17 \\
        Qwen2.5-VL-3B-SFT         & 1000 steps & 15.23 & 12.26 & 16.11 & 16.41 & 12.60 & 17.54 \\
        Qwen2.5-VL-3B-SFT         & 1 epoch & 16.77 & 14.20 & 17.53 & 18.05 & 14.45 & 19.13 \\
        Qwen2.5-VL-3B-GRPO        & 400 steps & 12.22 & 14.56 & 11.52 & 13.12 & 14.84 & 12.60 \\
        Qwen2.5-VL-3B-GRPO        & 1000 steps & 13.25 & 15.22 & 12.66 & 14.18 & 15.60 & 13.75 \\
        Qwen2.5-VL-3B-GRPO        & 1 epoch & 14.65 & 15.90 & 14.28 & 15.48 & 16.14 & 15.28 \\
        Qwen3-VL-4B-SFT           & 1 epoch & 22.46 & 21.53 & 22.73 & 24.04 & 21.97 & 24.66 \\
        Rex-Omni-3B~\cite{jiang2025detect} (HOI-SFT) & 1 epoch & \textbf{25.30} & 21.24 & \textbf{26.51} & \underline{26.97} & 21.79 & \textbf{28.51} \\
        \midrule
        \multicolumn{8}{@{}l}{\textit{--- HOI-R1 (SFT+GRPO) ---}} \\
        \addlinespace
        HOI-R1-Qwen2.5-VL-3B    & 1 epoch + 40 steps & 18.33 & 16.03 & 19.02 & 19.83 & 16.25 & 20.90 \\
        HOI-R1-Qwen3-VL-4B      & 1 epoch + 40 steps & 23.35 & \textbf{23.58} & 23.28 & 25.12 & \textbf{24.11} & 25.42 \\
        HOI-R1-Rex-Omni-3B      & 1 epoch + 40 steps & \underline{25.24} & \underline{22.57} & \underline{26.04} & \textbf{27.00} & \underline{23.21} & \underline{28.14} \\
        \bottomrule
    \end{tabular}
    }
    \vspace{1mm}
    \caption{Comparison on the HICO-DET dataset.}
    \label{tab:main_results_summary}
\end{table}
\begin{table}[!ht]
    \begin{minipage}{0.49\linewidth}
        \centering
        \resizebox{\linewidth}{!}{
            \begin{tabular}{@{}cccccccc@{}}
                \toprule
                \multicolumn{2}{c}{Settings} & \multicolumn{3}{c}{Default} & \multicolumn{3}{c}{Known Object} \\
                \cmidrule(lr){1-2} \cmidrule(lr){3-5} \cmidrule(lr){6-8}
                Thinking & Task Description & Full & Rare & Non-Rare & Full & Rare & Non-Rare \\
                \midrule
                \ding{51} & \ding{51} & \textbf{8.39} & \textbf{9.60} & \textbf{8.03} & \textbf{8.96} & \textbf{9.83} & \textbf{8.70} \\
                & \ding{51}   & 7.91 & 8.30 & 7.79 & 8.60 & 8.60 & 8.60 \\
                \ding{51} &  & 3.06 & 2.82 & 3.13 & 3.11 & 2.82 & 3.20 \\
                &  & 2.75 & 1.86 & 3.13 & 2.88 & 1.86 & 3.19 \\
                \bottomrule
            \end{tabular}
        }
        \vspace{1mm}
        \caption{Ablation studies of prompt design. The original Qwen2.5-VL-3B-Instruct model is used.}
        \label{tab:ablation_prompt}
    \end{minipage}
    \hfill
    \begin{minipage}{0.49\linewidth}
        \centering
        \vspace{2.5mm}
        \resizebox{\linewidth}{!}{
            \begin{tabular}{@{}ccccccc@{}}
                \toprule
                \multirow{2}{*}{Methods} & \multicolumn{3}{c}{Default} & \multicolumn{3}{c}{Known Object} \\
                \cmidrule(lr){2-4} \cmidrule(lr){5-7}
                Full & Rare & Non-Rare & Full & Rare & Non-Rare \\
                \midrule
                HOI-R1 & \textbf{13.25} & \textbf{15.22} & \textbf{12.66} & \textbf{14.18} & \textbf{15.60} & \textbf{13.75} \\
                w/o label reward & 12.49 & 14.78 & 11.80 & 13.54 & 15.26 & 13.02 \\
                w/o IoU reward   & 9.63  & 11.12 & 9.19  & 10.29 & 11.38 & 9.96  \\
                \bottomrule
            \end{tabular}
        }
        \vspace{1mm}
        \caption{Ablation studies of reward functions. All of the Qwen2.5-VL-3B-Instruct models are trained for 1,000 steps by GRPO.}
        \label{tab:ablation_reward}
    \end{minipage}
\end{table}

\subsection{Experimental Settings}
\noindent{\textbf{Dataset and Metric.}}\quad We conduct experiments on the HICO-DET~\cite{chao2018learning} dataset, a widely used benchmark for Human-Object Interaction (HOI) detection.
The dataset consists of 38,118 training images and 9,658 test images, encompassing 600 HOI categories formed by 117 verbs and 80 objects.
The HOI categories are further divided into three subsets based on the number of instances: \textit{Full}, \textit{Rare}, and \textit{Non-Rare}.
In addition, the evaluation is split into two settings: \textit{Default} and \textit{Known Object}, where the latter do not include unknown objects.
The mean Average Precision (mAP) is employed as the primary evaluation metric, calculated using an IoU threshold of 0.5 for both human and object bounding boxes, and the object and verb label must be correctly predicted.

\subsection{Implementation Details}
Our main experiments are conducted on the Qwen2.5-VL-3B-Instruct~\cite{bai2025qwen2} model, and we also validate the effectiveness of our method on the Qwen3-VL-4B-Instruct~\cite{Qwen3-VL} and Rex-Omni-3B~\cite{jiang2025rexomni} models.
In SFT stage, the GPT4o-mini~\cite{achiam2023gpt} is used to generate the thinking process for each training image in the HICO-DET dataset. The ground-truth annotations are converted into our desired output format. For an HOI pair with multiple interactions, we merge them into a single dict entry with a list of verbs. The full model is trained for 1 epoch with a batch size of 8. The AdamW optimizer~\cite{loshchilov2018decoupled} is used with a learning rate of 1e-6, and a cosine learning rate scheduler is applied.
Next, in the RL stage, the model is trained for 40 steps with a batch size of 16. The group size $G$ for GRPO is set to 4. The AdamW optimizer is used with a learning rate of 1e-6, and a linear learning rate scheduler is applied.
In the reward functions, the weights for the HOI key format reward are all set to $0.2$.

\subsection{Comparison to Baselines and HOID Methods}

In Table~\ref{tab:main_results_summary}, we compare our proposed HOI-R1 method with traditional HOID methods and MLLMs with different scales.
First, we evaluate the original Qwen-VL models with our designed HOID prompt in a training-free manner.
As the result, our baseline models, Qwen2.5-VL-3B and Qwen3-VL-4B achieves 8.39 and 16.16 mAP on the \textit{Full} category under the \textit{Default} setting, which is higher than the traditional HOID method HO-RCNN~\cite{chao2018learning}.
We also find that the performance on the \textit{Rare} category is better than that on the \textit{Full} and \textit{Non-Rare} categories, which is different from traditional HOID methods.
We consider the reason is that the image contains rare HOI categories usually have fewer HOI instances, making it easier for the MLLM to focus on the relevant interactions and utilize its strong reasoning ability to identify them.
Especially, the 32B and 72B model outperform the transformer-based HOID method, QPIC~\cite{tamura2021qpic}, demonstrating the strong prior knowledge of large MLLMs.

Next, we conduct SFT or RL training on the baseline model.
From the results of Qwen2.5-V-3B 400-step training, GRPO outperforms SFT by 0.51 mAP, and also from the training curve in Fig.~\ref{fig:results_all}, RL training increase the performance more effectively than SFT in the early training stage.
As our SFT training introduces HOI-specific knowledge from a teacher model, after 1 epoch training, with additional prior knowledge, the SFT model achieves 16.77 mAP, which is 2x higher than the baseline model.
For the Rex-Omni model which does not have reasoning ability, we do SFT training without the thinking process part in the prompt.
The Rex-Omni-HOI-SFT model achieves 25.30 mAP on the \textit{Full} category, which is higher than HOTR~\cite{kim2021hotr}.
Notably, our HOI-R1 model only train for 1 epoch, which converges much faster than traditional HOID methods that require training for hundreds of epochs.
The SFT training increases task specific knowledge, and the RL training enhances the generalization capabilities.
Our HOI-R1 combines both the benefits of the two training stages, Rex-Omni-3B model achieves 27.00 mAP
on the \textit{Full} categories under the \textit{Known Object} setting, which is 17.7\% higher than Qwen2.5-VL-72B-AWQ, and Qwen3-VL-4B achieves 23.58 mAP on the \textit{Rare} categories, which is 7.9\% higher than QPIC~\cite{tamura2021qpic}.

\begin{figure}[!t]
    \centering
    \includegraphics[width=0.85\linewidth]{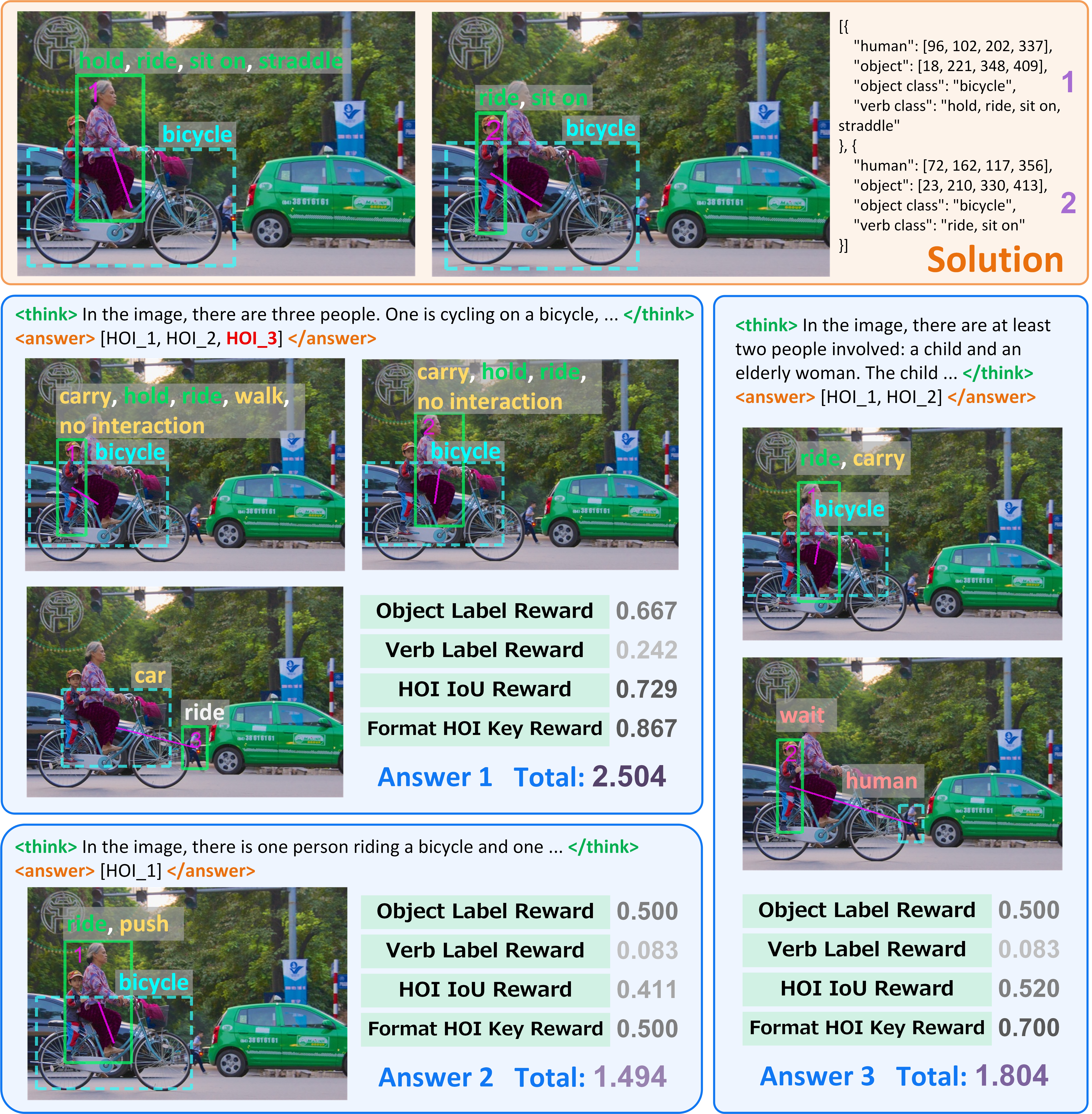}
    \caption{An example of the reward advantage of GRPO. The predicted HOI instances are visualized on the left with the rewards on the right.}
    \label{fig:reward_group}
\end{figure}

\subsection{Ablation Study}

\noindent{\textbf{Reward Functions.}}\quad To isolate and evaluate the effectiveness of our prompt design, in Table~\ref{tab:ablation_prompt}, we conduct the ablation study on the original Qwen2.5-VL-3B baseline model.
Comparing line 1 and line 2 and line 3 and line 4, we find that removing the chain-of-thought part mainly decreases the performance on the \textit{Rare} categories, indicating that the thinking process helps the model to reason about less common interactions.
Moreover, removing the task description part significantly degrades the performance on all categories, demonstrating that providing clear instructions is crucial for the model to understand and perform the HOID task effectively.

\noindent{\textbf{Reward Functions.}}\quad Our reward functions are designed specifically for the HOID task. To evaluate their effectiveness, we conduct drop-one experiments on our full model in Table~\ref{tab:ablation_reward}.
From the result, removing the label reward results in a decrease of 0.76 mAP on the \textit{Full} category under the \textit{Default} setting, indicating that the label reward, which encourages correct verb and object predictions, is crucial for improving the model's performance.
Furthermore, removing the IoU reward leads to a more significant drop of 3.62 mAP, highlighting the importance of accurate localization in HOID tasks.
These results demonstrate that each component of our reward functions contributes to the overall performance, and their combination is essential for achieving optimal results.

\subsection{Qualitative Results}
In Fig.~\ref{fig:reward_group}, we present a visualization results of a group of reward case. From the result, answer 1 has a higher reward than answer 2, as it correctly identifies more human-object interactions with accurate bounding boxes, while answer 3 misidentifies the interaction as ``wait'' and has less precise bounding boxes. The reward reflects these differences, showing that our designed reward functions effectively guide the model to generate more accurate HOI predictions. This example highlights the advantage of our GRPO training in enhancing the model's ability to accurately detect human-object interactions. 

\section{Conclusion}
\label{sec:conclusion}
In this paper, we present HOI-R1, the first pure MLLM framework for HOID tasks, which eliminates the need for object detectors. 
We introduce a novel two-stage training paradigm that combines supervised fine-tuning (SFT) and reinforcement learning (RL) to effectively adapt MLLMs for HOID tasks.
The SFT stage focuses on enhancing the model's ability to recognize human-object interactions through carefully designed instruction templates and data augmentation techniques.
The RL stage further refines the model's performance by optimizing it for specific HOID metrics, ensuring that the model not only understands the interactions but also excels in practical evaluation scenarios.
With our proposed SFT and RL paradigm, HOI-R1 achieves a significant performance boost on the HICO-DET dataset. Our results demonstrate the potential of MLLMs in structured tasks like HOID, paving the way for future research in this direction.


%
%
%
\bibliographystyle{splncs04}
\bibliography{main}

@String(CVPR= {IEEE Conf. Comput. Vis. Pattern Recog.})

@String(NIPS= {Adv. Neural Inform. Process. Syst.})

@String(ICML = {ICML})

@String(ICLR = {Int. Conf. Learn. Represent.})

@String(CVPR  = {CVPR})

@String(NIPS  = {NeurIPS})

@String(ICML  = {ICML})

@String(ICLR  = {ICLR})

@String(WACV = {WACV})

@inproceedings{chao2018learning,
  title={Learning to detect human-object interactions},
  author={Chao, Yu-Wei and Liu, Yunfan and Liu, Xieyang and Zeng, Huayi and Deng, Jia},
  booktitle=WACV,
  year={2018}
}

@inproceedings{kim2021hotr,
  title={{HOTR}: End-to-end human-object interaction detection with transformers},
  author={Kim, Bumsoo and Lee, Junhyun and Kang, Jaewoo and Kim, Eun-Sol and Kim, Hyunwoo J},
  booktitle=CVPR,
  year={2021}
}

@inproceedings{zou2021end,
  title={End-to-end human object interaction detection with hoi transformer},
  author={Zou, Cheng and Wang, Bohan and Hu, Yue and Liu, Junqi and Wu, Qian and Zhao, Yu and Li, Boxun and Zhang, Chenguang and Zhang, Chi and Wei, Yichen and others},
  booktitle=CVPR,
  year={2021}
}

@inproceedings{tamura2021qpic,
  title={{QPIC}: Query-based pairwise human-object interaction detection with image-wide contextual information},
  author={Tamura, Masato and Ohashi, Hiroki and Yoshinaga, Tomoaki},
  booktitle=CVPR,
  year={2021}
}

@article{kuhn1955hungarian,
  title={The Hungarian method for the assignment problem},
  author={Kuhn, Harold W},
  journal={Naval Res. Logist. Quart},
  pages={83--97},
  year={1955},
}

@inproceedings{loshchilov2018decoupled,
  title={Decoupled Weight Decay Regularization},
  author={Loshchilov, Ilya and Hutter, Frank},
  booktitle=ICLR,
  year={2018}
}

@inproceedings{radford2021learning,
  title={Learning Transferable Visual Models From Natural Language Supervision},
  author={Radford, Alec and Kim, Jong Wook and Hallacy, Chris and Ramesh, Aditya and Goh, Gabriel and Agarwal, Sandhini and Sastry, Girish and Askell, Amanda and Mishkin, Pamela and Clark, Jack and others},
  booktitle = ICML,
  year={2021}
}

@inproceedings{ning2023hoiclip,
  title={HOICLIP: Efficient Knowledge Transfer for HOI Detection with Vision-Language Models},
  author={Ning, Shan and Qiu, Longtian and Liu, Yongfei and He, Xuming},
  booktitle=CVPR,
  year={2023}
}

@article{cao2024detecting,
  title={Detecting any human-object interaction relationship: Universal hoi detector with spatial prompt learning on foundation models},
  author={Cao, Yichao and Tang, Qingfei and Su, Xiu and Chen, Song and You, Shan and Lu, Xiaobo and Xu, Chang},
  booktitle=NIPS,
  year={2024}
}

@inproceedings{li2023blip,
  title={Blip-2: Bootstrapping language-image pre-training with frozen image encoders and large language models},
  author={Li, Junnan and Li, Dongxu and Savarese, Silvio and Hoi, Steven},
  booktitle={International conference on machine learning},
  pages={19730--19742},
  year={2023},
  organization={PMLR}
}

@inproceedings{chen2025focusing,
  title={Focusing on what to Decode and what to Train: SOV Decoding with Specific Target Guided DeNoising and Vision Language Advisor},
  author={Chen, Junwen and Wang, Yingcheng and Yanai, Keiji},
  booktitle=WACV,
  year={2025}
}

@article{shen2025vlm,
  title={Vlm-r1: A stable and generalizable r1-style large vision-language model},
  author={Shen, Haozhan and Liu, Peng and Li, Jingcheng and Fang, Chunxin and Ma, Yibo and Liao, Jiajia and Shen, Qiaoli and Zhang, Zilun and Zhao, Kangjia and Zhang, Qianqian and others},
  journal={arXiv preprint arXiv:2504.07615},
  year={2025}
}

@article{bai2025qwen2,
  title={Qwen2. 5-vl technical report},
  author={Bai, Shuai and Chen, Keqin and Liu, Xuejing and Wang, Jialin and Ge, Wenbin and Song, Sibo and Dang, Kai and Wang, Peng and Wang, Shijie and Tang, Jun and others},
  journal={arXiv preprint arXiv:2502.13923},
  year={2025}
}

@article{Qwen3-VL,
  title={Qwen3-VL Technical Report}, 
  author={Shuai Bai and Yuxuan Cai and Ruizhe Chen and Keqin Chen and Xionghui Chen and Zesen Cheng and Lianghao Deng and Wei Ding and Chang Gao and Chunjiang Ge and others},
  journal={arXiv preprint arXiv:2511.21631},
  year={2025}
}

@article{jiang2025rexomni,
      title={Detect Anything via Next Point Prediction}, 
      author={Qing Jiang and Junan Huo and Xingyu Chen and Yuda Xiong and Zhaoyang Zeng and Yihao Chen and Tianhe Ren and Junzhi Yu and Lei Zhang},
      year={2025},
      journal={arXiv preprint arXiv:2510.12798}
}

@article{guo2025deepseek,
  title={Deepseek-r1: Incentivizing reasoning capability in llms via reinforcement learning},
  author={Guo, Daya and Yang, Dejian and Zhang, Haowei and Song, Junxiao and Zhang, Ruoyu and Xu, Runxin and Zhu, Qihao and Ma, Shirong and Wang, Peiyi and Bi, Xiao and others},
  journal={arXiv preprint arXiv:2501.12948},
  year={2025}
}

@article{zhu2025internvl3,
  title={Internvl3: Exploring advanced training and test-time recipes for open-source multimodal models},
  author={Zhu, Jinguo and Wang, Weiyun and Chen, Zhe and Liu, Zhaoyang and Ye, Shenglong and Gu, Lixin and Tian, Hao and Duan, Yuchen and Su, Weijie and Shao, Jie and others},
  journal={arXiv preprint arXiv:2504.10479},
  year={2025}
}

@article{li2024llava,
  title={Llava-onevision: Easy visual task transfer},
  author={Li, Bo and Zhang, Yuanhan and Guo, Dong and Zhang, Renrui and Li, Feng and Zhang, Hao and Zhang, Kaichen and Zhang, Peiyuan and Li, Yanwei and Liu, Ziwei and others},
  journal={arXiv preprint arXiv:2408.03326},
  year={2024}
}

@article{huang2025vision,
  title={Vision-r1: Incentivizing reasoning capability in multimodal large language models},
  author={Huang, Wenxuan and Jia, Bohan and Zhai, Zijie and Cao, Shaosheng and Ye, Zheyu and Zhao, Fei and Xu, Zhe and Hu, Yao and Lin, Shaohui},
  journal={arXiv preprint arXiv:2503.06749},
  year={2025}
}

@inproceedings{tan2025reason,
  title={Reason-rft: Reinforcement fine-tuning for visual reasoning of vision language models},
  author={Tan, Huajie and Ji, Yuheng and Hao, Xiaoshuai and Chen, Xiansheng and Wang, Pengwei and Wang, Zhongyuan and Zhang, Shanghang},
  booktitle=NIPS,
  year={2025}
}

@article{shao2024deepseekmath,
  title={Deepseekmath: Pushing the limits of mathematical reasoning in open language models},
  author={Shao, Zhihong and Wang, Peiyi and Zhu, Qihao and Xu, Runxin and Song, Junxiao and Bi, Xiao and Zhang, Haowei and Zhang, Mingchuan and Li, YK and Wu, Yang and others},
  journal={arXiv preprint arXiv:2402.03300},
  year={2024}
}

@article{achiam2023gpt,
  title={Gpt-4 technical report},
  author={Achiam, Josh and Adler, Steven and Agarwal, Sandhini and Ahmad, Lama and Akkaya, Ilge and Aleman, Florencia Leoni and Almeida, Diogo and Altenschmidt, Janko and Altman, Sam and Anadkat, Shyamal and others},
  journal={arXiv preprint arXiv:2303.08774},
  year={2023}
}

@article{jiang2025detect,
  title={Detect Anything via Next Point Prediction},
  author={Jiang, Qing and Huo, Junan and Chen, Xingyu and Xiong, Yuda and Zeng, Zhaoyang and Chen, Yihao and Ren, Tianhe and Yu, Junzhi and Zhang, Lei},
  journal={arXiv preprint arXiv:2510.12798},
  year={2025}
}
\end{document}